\documentclass[journal]{IEEEtran}
\usepackage{times}
\usepackage{epsfig}
\usepackage{graphicx}
\usepackage{amsmath}
\usepackage{amssymb}

\usepackage{comment}
\usepackage{cuted}
\usepackage{capt-of}
\usepackage[switch]{lineno}
\usepackage{cite}
\usepackage{amsmath}
\usepackage{array}
\begin{document}
%
\title{Hair Segmentation on Time-of-Flight RGBD Images}
%
%
%

\author{Yuanxi Ma, Cen Wang, Shiying Li, Jingyi Yu
\thanks{All authors are with the School of Information Science and Technology, ShanghaiTech University, 393 Middle Huaxia Road, Pudong, Shanghai, China.}}
\maketitle

\begin{abstract}
   Robust segmentation of hair from portrait images remains challenging: hair does not conform to a uniform shape, style or even color; dark hair in particular lacks features. We present a novel computational imaging solution that tackles the problem from both input and processing fronts. We explore using Time-of-Flight (ToF) RGBD sensors on recent mobile devices. We first conduct a comprehensive analysis to show that scattering and inter-reflection cause different noise patterns on hair vs. non-hair regions on ToF images, by changing the light path and/or combining multiple paths. We then develop a deep network based approach that employs both ToF depth map and RGB image to produce an initial hair segmentation with labeled hair components. We then refine the result by imposing ToF noise prior under the conditional random field. We collect the first ToF RGBD hair dataset with 20k+ head images captured on 30 human subjects with a variety of hairstyles at different view angles. Comprehensive experiments show that our approach outperforms the state-of-the-art techniques in accuracy and robustness and can handle traditionally challenging cases such as dark hair, similar hair/background, similar hair/foreground, etc. 
\end{abstract}


%
\IEEEpeerreviewmaketitle

\section{Introduction}
%
%
%
%
\IEEEPARstart{H}{air} extraction and modeling play an important role in human identification in computer vision as well as in character creation for digital entertainment production. Robust localization of the hair component and accurate recognition of hair styles in face images can improve face tracking and recognition; the segmented hair component can be separately modeled to create realistic 3D human avatars. By far nearly all existing approaches operate on RGB images by treating hair as a special semantic label. In reality, hair does not conform to a uniform shape, style or even color; black hair in particular lacks features, making hair segmentation in the wild an extremely difficult task.
\begin{figure}
    \centering
    \includegraphics[width=1\linewidth]{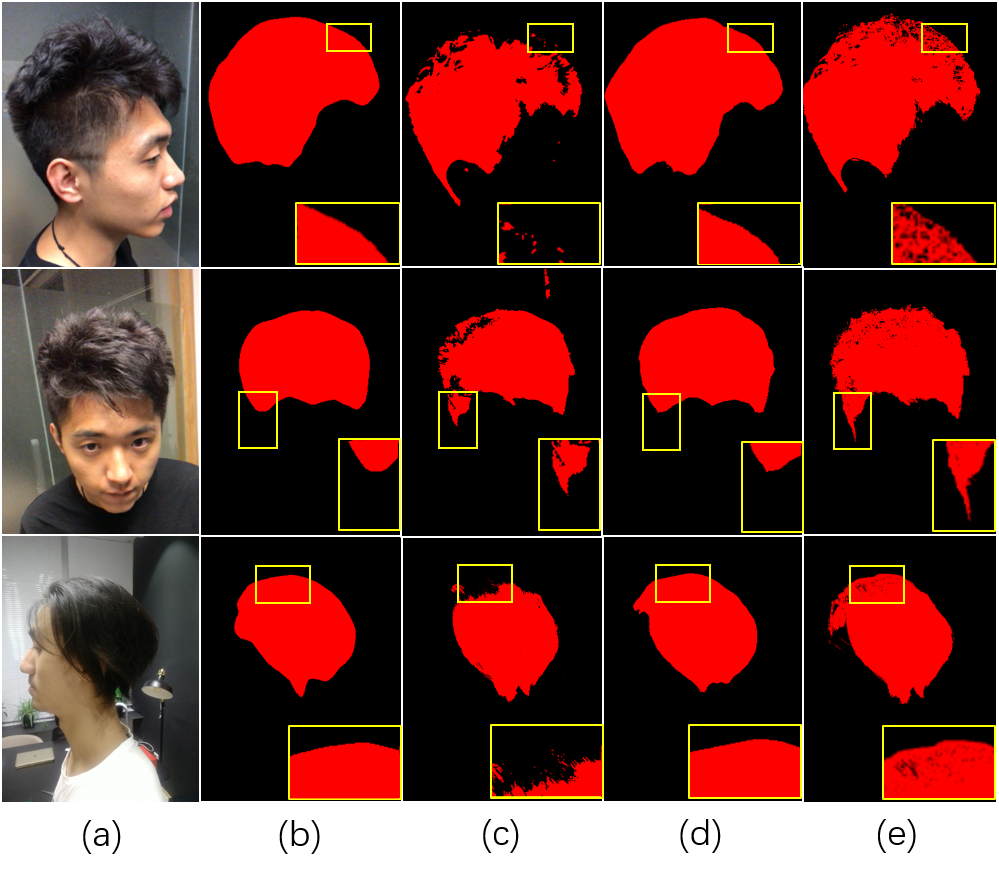}
    \caption{Hair segmentation in challenging scenes. Rows: similar appearance to the background, hair at the temples and highlights on hair. Columns: (a) RGB input, and hair masks output from (b) D-CNN \cite{wang2018depth}, (c) CRFs \cite{krahenbuhl2011efficient}, (d) DeepLabv3+ \cite{chen2018encoder} and (e) Our method. }  
    \label{fig:tensor}  
\end{figure}

Computer graphics approaches resort to human interventions: a user manually provides a coarse segmentation followed by automatic refinement \cite{c2014braid}. For example, several recent methods allow the user to draw a small number of strokes in the hair regions \cite{2013TOGchai, Yu2014, m2017TOG} where features on these strokes are then used to guide the hair segmentation of the whole image. Hand-crafted features such as histogram of oriented gradients and local ternary patterns \cite{2014AAAI, 2016ICIP} have shown better performance than the classical ones such as SIFT. Fully automated methods have sought to employ machine learning techniques by training on a large number of portrait images with manually annotated hair regions \cite{k2016auto, m2018ECCV, liang2018video}. These approaches follow the general segmentation pipeline, e.g., the pyramid scene parsing network (PSPNet) \cite{2017CVPRzhao}. However, hair is far away from being a "normal" semantic label: it is semitransparent and has variable shapes. Even using a large set of training data, automatic approaches are vulnerable to lighting, viewing angles, background similarities, etc. Figure \ref{fig:tensor} shows common failure cases using three state-of-the-art segmentation techniques: depth-aware convolutional neural networks (D-CNN for short) \cite{wang2018depth}, the fully connected conditional random fields (CRFs) \cite{krahenbuhl2011efficient} and DeepLabv3+ \cite{chen2018encoder}. 

In this paper, we present a computational photography approach that tackles the problem from the input front. We explore hair segmentation on RGBD images captured by time-of-flight (ToF) cameras on mobile devices. Several latest mobile phones are equipped with ToF: OPPO r17 pro uses a $240\times180$ ToF sensor whereas VIVO x20 uses an even higher resolution sensor of $640\times480$. We first conduct a comprehensive analysis to show that hair scattering and inter-reflection cause different noise patterns on hair vs. non-hair regions on ToF sensors. Specifically scattering causes the multi-path artifacts whereas inter-reflection changes the length of light paths. Since hair strands exhibit randomness in spatial arrangement, their corresponding scattering and inter-reflection pattern follow similar distributions. 

For verification and subsequently training, we collect the first RGBD hair dataset with 20k+ head images captured on 30 subjects with a variety of hairstyles at different view angles. We simultaneously annotate hair in terms of top, back and two side regions, and we show that the noise in ToF depth images is coherent across these regions but is missing on skins. At the same time, the gradient fields in RGB images exhibit heterogeneity across different regions. We therefore develop a deep network approach that employs both the ToF noise characteristics and the gradient fields on RGB images. We adopt the DeepLabv3+ \cite{chen2018encoder} framework and modify the encoder/decoder architecture to provide an initial segmentation with region labels. We further refine the results using the conditional random field (CRFs) with the ToF noise model as a prior. Comprehensive experiments show that our approach significantly outperforms the state-of-the-art RGB based techniques in accuracy and robustness. In particular, our technique can handle challenging segmentation cases such as dark hair, similar hair/background, similar hair/foreground, hair viewed at different angles, etc. In addition, we show that our labeled hair segmentation can be further used to identify hair styles.

\section{Related Work}
Hair is a most challenging type of objects in recognition, segmentation, and reconstruction since hair is translucent with filamentous structure. Hair modeling and reconstruction aims to produce lifelike hair for virtual human in game and film production as well as to beautify portraits for hairstyle tryon. Image-based approaches achieve higher quality with less efforts than physical simulation based methods (see \cite{2007TVCG} and \cite{m2018Acess} for a comprehensive survey). The core of the problem lies in how to segment the hair component in images. 

Conventional approaches \cite{2014AAAI, 2016ICIP} extract hair features from texture, shape and color, and then exploit machine learning techniques such as random forests \cite{2001Breiman} and support vector networks \cite{1995Cortes} to separate the hair vs. non-hair regions. In general, light colored hair is easier to segment than the dark one as it exhibits richer features. The accuracy of these approaches depends heavily on the difference between the foreground and the background. Semi-automatic approaches allow the users to draw strokes or splines on the hair regions \cite{2013TOGchai, Yu2014, 2016TOGreal, m2017TOG} or generate seed pixels \cite{2013CVPRsmith} to produce more accurate hair boundaries, remove outlier regions and reduce computational expense. 

More recent hair segmentation techniques employ deep convolutional neural networks (CNN) by learning to produce more discriminative features even better than the hand-crafted ones. Approaches in this category generally train on a large number of manually annotated portrait image datasets and employ semantic segmentation pipelines such as PSPNet \cite{2017CVPRzhao} to automatically obtain pixel-wise hair masks. \cite{k2015CVPR} adopts a multi-objective CNN for both pixelwise labels and a pairwise edge map. \cite{k2016auto} applies Region-CNN (R-CNN) to estimate hair distribution classes and then generate a hair mask with a directional map. \cite{k2017ICME} and \cite{liang2018video} show that fully convolutional networks (FCN) achieve higher accuracy in hair segmentation. By far the most approaches use RGB color features whereas we exploit the depth channel, more precisely, the noise pattern on the depth channel. 

Our work also seeks to automatically determine hair styles, an active research area in the field of computer graphics and vision. Most methods require user interventions to indicate the hair styles (e.g., via directional strokes) and then search for best matching examples in a large hairstyle database manually constructed from public online repositories \cite{2018Arts}. They optimize the discrepancy between the reference photo and the selected hairstyle in the database, and synthesize the hair strands to fit to the hairstyle \cite{m2015SIGGRAPHhu, m2017TOG}. We in contrast show that ToF + RGB images can be used to automatically infer hair styles, by simultaneously employing color, gradient, and noise patterns.

\section{Time-of-Flight Image Analysis on Hair}
\label{section: noise analysis}
A Time-of-Flight (ToF) camera works in conjunction with an active illumination modulated in intensity. Light emitting from the camera is reflected at the surface of an object and then sensed by a pixel array in the camera. The received ray attenuates according to the surface reflectance and its travel distance. Often, a narrow-band filter is used on ToF to prevent ambient light interference.

To measure depth, one can compute the phase shift that light travels from the light source to the object surface and then back to the sensor. We follow the same notation as \cite{heide2013low,su2016material} where a correlation pixel measures the modulated exposure as:
\begin{figure}
	\centering
	\includegraphics[width=1\linewidth]{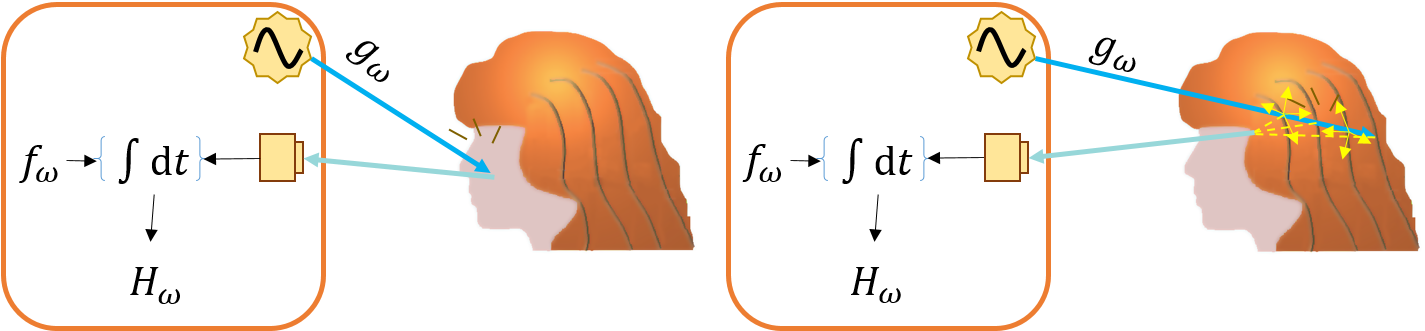}
	\caption{A ToF camera captures light along a single path reflected at the skin (left) or longer and/or multiple light paths at hair due to scattering and inter-reflection (right). }  
	\label{fig:tof}  
\end{figure}

\begin{equation}
H_{\omega ,\phi }=\int _{0}^{T}E(t)f_{\omega}(t+\phi / \omega)dt
\label{eq1}
\end{equation}

\noindent where $E(t)$ is irradiance, $f_{\omega}$ is a periodic reference function with modulation period $T$ of angular frequency $\omega$ and programmable phase offset $\phi$, both evaluated at time $t$. The reference function is generally a zero-mean periodic function such as sinusoidal and rectangle waves.

\subsection{Skin vs. Hair}
Compared with the smooth skin regions of face and neck, hair regions are volumetric, filamentous and textured. Light arriving at the surface of hair is scattered in the filaments repetitively and reflected to the sensor, as shown in Figure \ref{fig:tof}. When a single light path from a smooth surface contributes to a sensor pixel, the irradiance at the pixel is:
\begin{equation}
E_{\omega}(t)=E_{0}+\alpha E_{m}g_{\omega}(t+\tau)
\label{eq2}
\end{equation}
where $E_0$ is the dark current component, $E_m$ the modulated amplitude of the light source. $\alpha$ presents an attenuation term, and $\tau$ the total travel time from the light source to the object surface and then to the sensor pixel. $g_\omega$ is intensity modulated at the same frequency as $f_\omega$.

We substitute $E(t)$ in Equation \ref{eq1} with $E_\omega(t)$, and obtain Equation \ref{eq3} for a smooth surface due to the zero-mean periodic function $f_\omega$.
\begin{equation}
\begin{split}
H_{smooth}=\int_{0}^{T}(E_{0}+\alpha  E_{m}g_{\omega}(t+\tau))f_{\omega}(t+\phi)dt \\=\alpha E_{m}\int_{0}^{T}g_{\omega}(t+\tau)f_{\omega}(t+\phi)dt
\end{split}
\label{eq3}
\end{equation}

When using ToF to image object surfaces of reflective and/or translucent materials, the inter-reflection and scattering can cause strong multi-path artifacts on the ToF image. However, by far such artifacts had been mostly used for modeling smooth surfaces, not uneven geometry such as hair as in this paper. For example, on the basis of the observation, \cite{su2016material} classifies structurally distinct materials using a high-precision ToF camera. In the case of hair, the multi-path artifacts are stronger: a light ray can be scattered by the hair surface, causing the light path to fork and subsequently the measured light path at a pixel is a combination of multiple light paths; a light ray can also be reflected by hair surface where dense hair fibers can significantly and randomly change the length of the path. 

We define the temporal (since ToF measures time/phase) point spread function $\alpha(\tau)$ to represent the integral of contributions from the light paths $p$ that correspond to the same travel time $\tau$, and yield Equation \ref{eq4} for a rough surface.
\begin{equation}
\begin{aligned}
&H_{rough}=E_{m}\int_{0}^{\infty } 
\alpha(\tau) \int_{0}^{T} g_{\omega}(t+\tau)f_{\omega}(t+\phi)dt d\tau,
\label{eq4}
\\
&\alpha(\tau)=\int_{P}(\delta (\left |  p\right |=\tau)\alpha_{p})dp
\end{aligned}
\end{equation}
where $\alpha_{p}$ is the light attenuation along a given light path $p$, which connects the light source to the sensor pixel. $P$ represents the space of all light paths, and $\left |  p\right |$ the travel time along path $p$.

Each pixel of the sensor hence receives the signal reflected at single or multiple light paths from different hair strands and produces a proportional electrical signal with the same frequency as the incident light. In practical, we can calculate the phase shift $\phi$ and amplitude $a$ of the sensed light using four equally spaced sampling points A1, A2, A3, and A4 per modulation period for a sinusoidal function \cite{lange2001solid} as\\
\begin{equation}
\begin{aligned}
&\phi=atan(\frac{A_{4}-A_{2}}{A_{1}-A_{3}}),\\ 
&a=\frac{ \sqrt{(A_{4}-A_{2})^{2}+(A_{1}-A_{3})^{2} }}{ 2 }\\
\end{aligned}
\end{equation}

Since  the phase shift is equivalent to time shift in a periodic signal, we can compute the travel distance of the light with the known light velocity c and modulation period T, as in Equation \ref{eq6}\\
\begin{equation}
d=\frac{c\ T\ \phi}{4\pi } \\
\label{eq6}
\end{equation}

\begin{figure}
	\centering
	\includegraphics[width=1\linewidth]{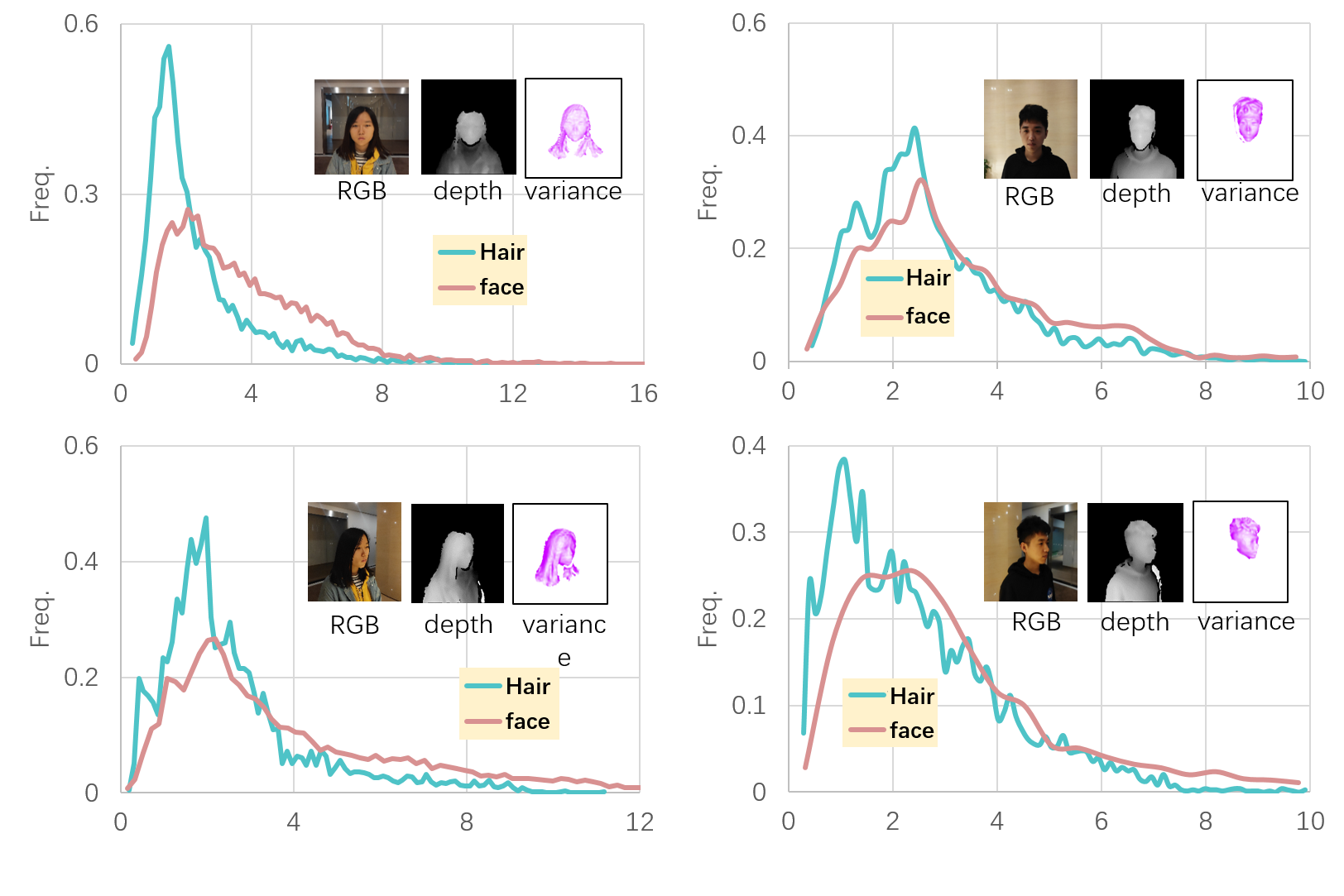}
	\caption{Variance map and its histogram curve of ToF at facial (red) vs. hair (blue) regions for different hair styles/regions.}  
	\label{fig:var}  
\end{figure}

Several recent studies have shown that in addition to reflection properties of object surface, distance between the object and the sensor and the background surrounding the object may also contribute to depth noise \cite{falie2007noise, sarbolandi2015kinect}. In our experiments we observe that inter-reflection and scattering artifacts impose much stronger noise than the ones caused by depth and background. Similar to \cite{falie2007noise, sarbolandi2015kinect}, we can then fit the depth noise using a Gaussian or a Poisson function, as shown in Figure \ref{fig:var}. In our analysis, we select a patch using a Gaussian window on the depth image, and fit it with a Gaussian function. The patch slides on the image so that we obtain a variance map. We calculate the variance map at the facial and hair regions for different subjects with long or short hairstyles, and also for the same subject at different viewing angles (top and side). We further consider different sizes of the selected patch for appropriate description of the variance map. For clarity, we show the variance histogram curves at the facial and hair regions of a subject, facial region in red and hair region in blue, respectively.

\begin{figure}
	\centering
	\includegraphics[width=1\linewidth]{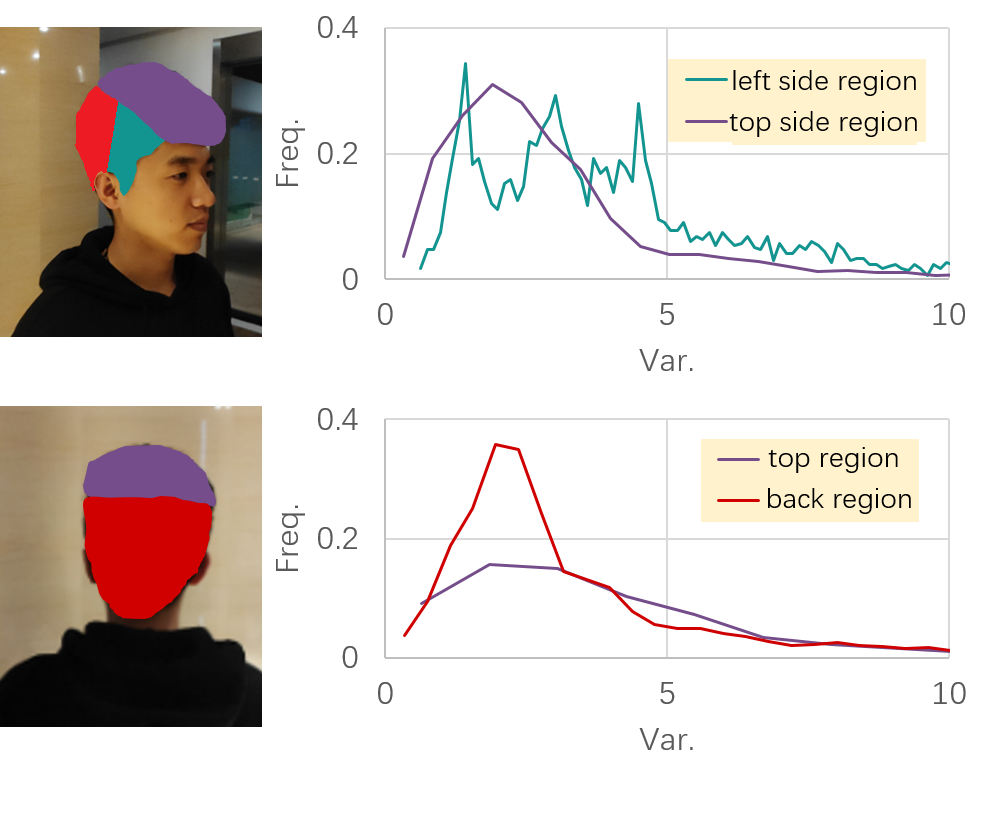}
	\caption{Variance histogram curve of ToP of the same hair captured from different views. }  
	\label{fig:var_part}  
\end{figure}

We observe that the variance histogram curve of hair region is highly distinguishable from that of facial region, and across different hairstyles. In experiments, we also observe that the variance histogram curves of hair are distinct at viewing angles of sides, back and top, as shown in Figure \ref{fig:var_part}. This is mainly because the hair density and directions exhibit strong variations at different view angles.

\subsection{ToF Hair Dataset}

\begin{figure}
	\centering
	\includegraphics[width=1\linewidth]{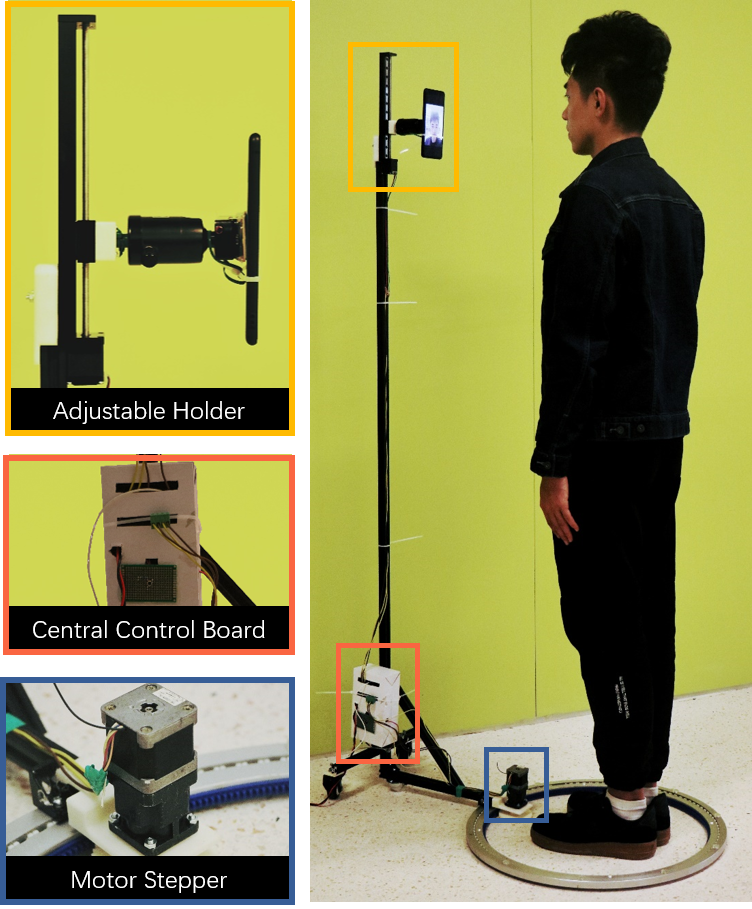}
	\caption{Our capture system and controllers.}  
 	\label{fig:dataset_2}  
\end{figure}

We collect the first ToF hair dataset for hair segmentation as shown in Figure \ref{fig:dataset_2}. Our capture system consists of a mobile phone, which is equipped with a high-precision RGB camera and a ToF depth camera, a gripper to hold the mobile phone, and a stepper with a controller. The phone held in the gripper orbits our subject, who stands immobile on the ground, with a radius of 1.5 meters, and take images of her/his head ranging from 0 to 360 degrees. For each subject, we select 720 images at the interval of 0.5 degrees. 

\begin{figure}
	\centering
	\includegraphics[width=1\linewidth]{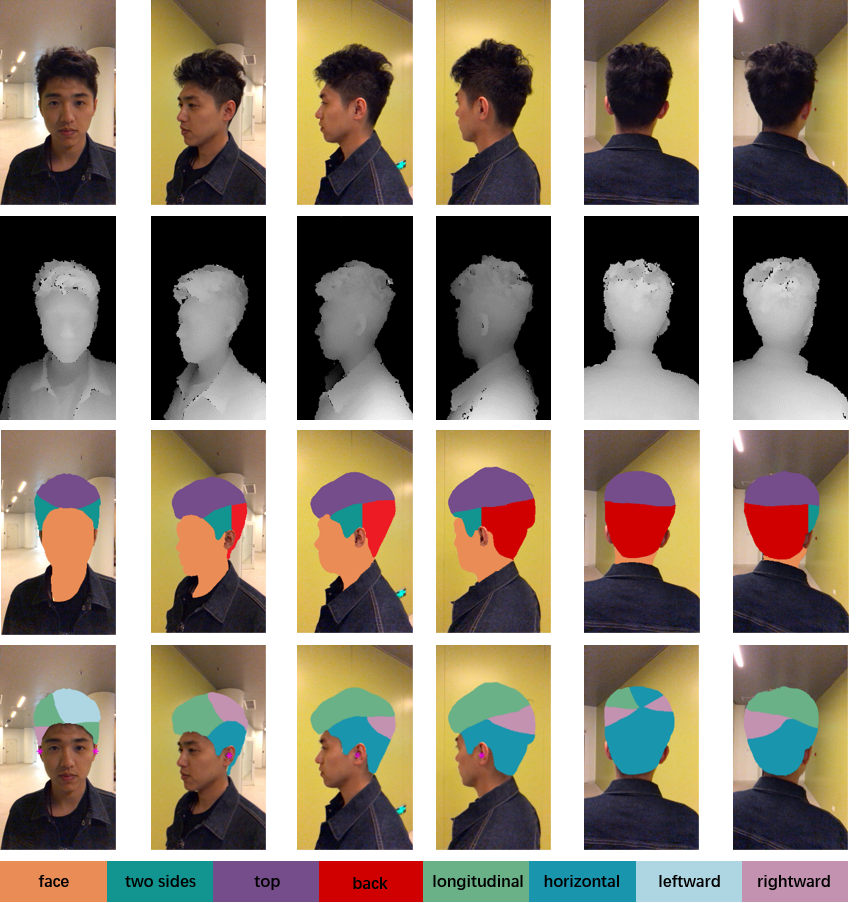}
	\caption{Image examples from our ToF hair dataset. Rows from top to bottom: RGB images, depth images, hair regions, directional maps. Columns from left to right: images captured at the interval of 45 degrees. }  
 	\label{fig:dataset_1}  
\end{figure}

Our dataset contains 20k+ head images of 30 subjects (18 males and 12 females), with various hairstyles and between the age of 18 and 40. We used the data of 23 subjects for training and the rest 7 for testing/validation. RGB images are 1024$\times$2048, and depth images 640$\times$480. We have five graphical artists to manually annotate pixelwise labels and take the average for ground-truth hair regions of top, two sides and back, and directional maps on the RGB images for training data, as shown in Figure \ref{fig:dataset_1}. In particular, we partition the hair regions of top along the curve above the eyebrows, and consider two ears as the landmark in locating the regions of back and two sides. For directional map annotation, we divide hair into four sections, i.e., $0^o-22.5^o$ and $157.5^o-180^o$ for horizontal, $80^o-110^o$ for longitudinal, $22.5^o-80^o$ for leftward, and $100^o-157.5^o$ for rightward. All angles are based on the image. We align the depth image with its corresponding RGB image using our alignment technique so that we have the hair regions and the directional map annotated on both images.


\section{Hair Segmentation and Labeling}
\label{section:network}
\begin{figure}
	\centering
	\includegraphics[width=1\linewidth]{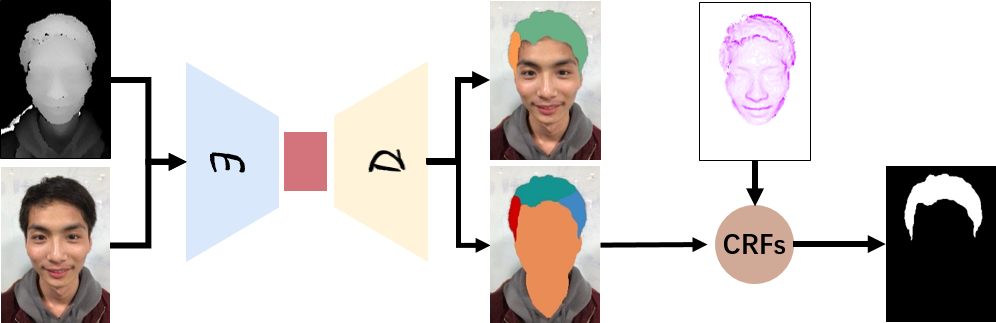}
	\caption{Our processing pipeline which combines a learning-based framework (ToF-HairNet) and CRFs based refinement.}  
	\label{fig:pipline}  
\end{figure}

We set out to use both the ToF depth map and the RGB images for conducting hair segmentation. A unique aspect of our approach is that we not only segment the hair region but also partition the region into semantically meaningful labels. \cite{gupta2014learning} demonstrates significant improvement in object segmentation by combining RGB images with three channels of HHA (Horizontal disparity, Height above ground and Angle that local normal makes with the inferred gravity direction) at each pixel computed from a depth image, instead of using raw depth in their CNN networks. \cite{wang2018depth} also present a depth-aware CNN for semantic segmentation of scenes on the basis of earlier version of Deeplabv3+ \cite{chen2018encoder}. According to its promising performance in \cite{chen2018encoder}, we exploit the DeepLabv3+ framework as the baseline of our network. Figure \ref{fig:pipline} shows our pipeline. The output of the network is the segmentation result that partitions the image into facial region (i.e., face and neck) and hair regions (top, back and two sides). We also conduct refinement to further improve the segmentation result. 
 
\subsection{Initial Segmentation}
\begin{figure*}
	\centering
	\includegraphics[width=1\textwidth]{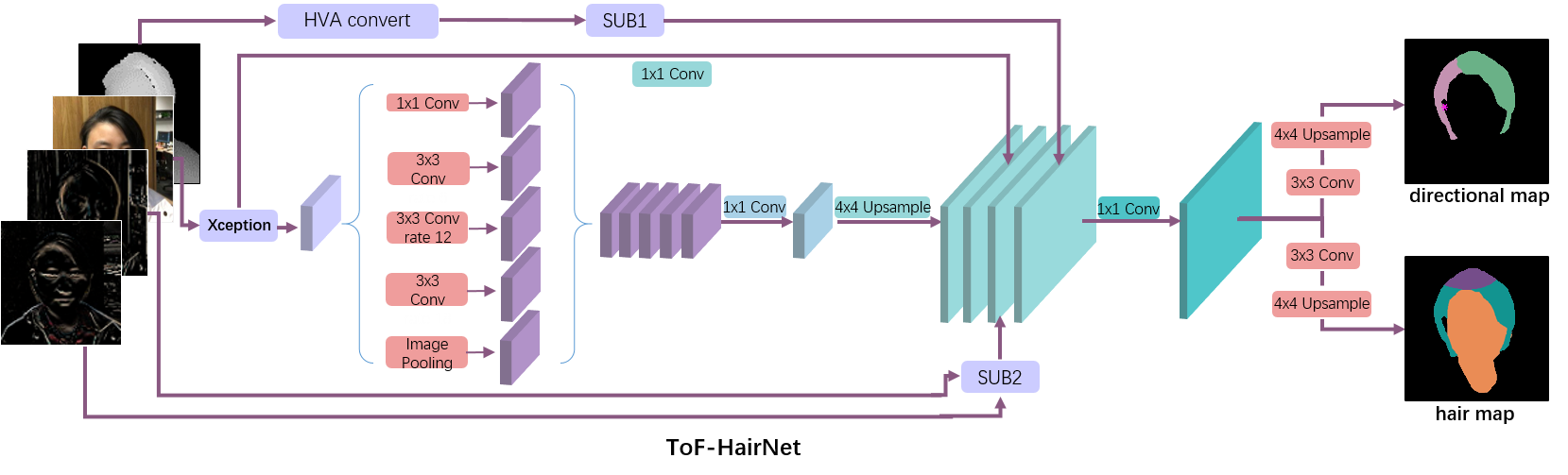}
	\caption{Our ToF-HairNet builds upon a 65-block Xception network as the backbone with a 80-layer network on the encoder side.  }  
	\label{fig:network}  
\end{figure*}

We construct our ToF-HairNet for multiple tasks based on the encoder and decoder of DeepLabv3+, with two subnets SUB1 and SUB2, as shown in Figure \ref{fig:network}. 

Recall that our RGB and ToF images have different sizes, 1024$\times$2048 and 640$\times$480, respectively. We first align the corresponding pair of RGB and ToF images based on the rotation-translation matrix provided by the mobile phone. The alignment yields to holes due to upsampling of the ToF image. We fill in these holes using a weighted mean value which we calculate from a patch of Gaussian window. We select the patch from its neighboring region annotated as the same label in its paired training data. In this way, our depth map (the upsampled ToF image) has its annotation and enables the computation of its variance map at specific regions. 

To calculate variance map, we extract the human region from the depth map. We select a patch $A$ of $7\times7$ in a Gaussian window $w_{i,j}$ to calculate its mean value $\bar{d}$ and variance $v_j$, and slide the patch on the image, as in Equation \ref{eq7}. 

\begin{equation}
v_{j}=\frac{\sum _{j\in A}w_{i,j}(d_{i}-\bar{d})^{2}}{N(A)}
\label{eq7}
\end{equation}

where $d_i$ is the value at pixel $i$ on the depth map, $N$ the number of pixels in patch $A$. We compute variance map on patch $A$ in sizes of $5\times5$, $7\times7$, $9\times9$ and $11\times11$, and exploit the average as the input variance map of the human region. From our analysis on ToF noise pattern (in \ref{section: noise analysis}), variance map at the facial region and hair regions show different properties. 

We then compute the two channels of horizontal disparity and angle that local normal makes with the inferred gravity direction from the depth map as in \cite{gupta2014learning}, and combine them with the variance map to obtain a map with three channels, which we call HVA channels for short.

\begin{figure}
	\centering
	\includegraphics[width=1\linewidth]{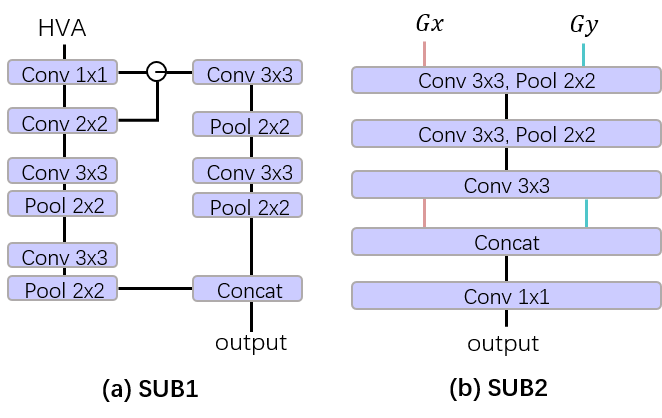}
	\caption{Two subnets of our ToF-HairNet: SUB1 and SUB2. }  
	\label{fig:subnets}  
\end{figure}

We feed HVA channels to the SUB1 (see Figure \ref{fig:subnets} (a)). We align the HVA channels with their corresponding RGB image with its background removed, to a 5-layer network of $3\times3$ convolution and a $1\times1$ convolution, and obtain features of 10 channels. In addition, we add the result at the second layer from the third layer to further fusion the low-level feature, then apply the residual to a 3-layer network of $3\times3$ convolution and a $1\times1$ convolution, and obtain features of 10 channels.

 Further, we observe that the direction of hair fibers provides important cues to both hair style and the location of the specific hair components. We therefore feed the RGB gradient maps as inputs to the SUB2 (see Figure \ref{fig:subnets} (b)). Specifically, we use a 65-block Xception network \cite{chollet2017xception} as our DCNN backbone to extract a high-level feature map from the RGB image to obtain features of 40 channels. We convert a RGB image to its gray image, calculate the gradients in $x$ and $y$ directions using a Sobel operator, and yield $G_x$ and $G_y$. We input $G_x$ and $G_y$ separately to a 5-layer network of $3\times3$ convolution for refinement. We concatenate the refined $G_x$ and $G_y$ into a gradient map and employ a $1\times1$ convolution to produce features of 20 channels. This process is shown in Figure \ref{fig:subnets} (b). The final network is composed of 80 layers, including 40 channels for a feature map from RGB images, 20 channels from HVA channels, and 20 channels from gradient maps, for fine-tuning the feature map at the encoder side.

On the decoder side, we adopt a $1\times1$ convolution to a low-level feature map which is extracted from the second block of the Xception network, and upsample the fine-tuned feature map by 4. We then concatenate the aligned ToF image, the gradient map, the low-level feature map and the fine-tuned feature map, and apply $1 \times 1$ convolution. Finally, after upsampling and a $1 \times 1$ convolution, our network outputs hair directional map and the segmentation result.

\subsection{Segmentation Refinement}

The initial segmentation output from our ToF-HairNet includes background, facial region, hair regions of top, back and two sides. Since hair exhibits strong appearance variations under different lighting and view angles, we further refine the segmentation results by employing dense conditional random fields plus extra features (CRFs+X for short).

The fully connected conditional random fields (CRFs)  in \cite{krahenbuhl2011efficient} represent a popular model for semantic segmentation and labeling. This model connects all pairs of individual pixels and define pairwise edge potentials by linearly combining Gaussian kernels. Following \cite{krahenbuhl2011efficient}, we consider a conditional random field as a Gibbs distribution. The Gibbs energy includes the unary potentials $\varphi _{u}(x_{i})$ and the pairwise potentials $\varphi _{p}(x_{i},x_{j})$, as in Equation \ref{eq8}:

\begin{equation}
E(x)=\sum _{i}\varphi _{u}(x_{i})+\sum_{i<j}\varphi _{p}(x_{i},x_{j})
\label{eq8}
\end{equation}

where i and j range from 1 to $N$. We compute the unary potential $\varphi _{u}(x_{i})$ independently for each pixel by a classifier that produces a distribution over the label assignment $x_{i}$ given image features. We normalize the network output and merge the different hair areas so that the output tensor only contain the hair area, face and background. The We utilize the output as the unary potential. The pairwise potentials $\varphi _{p}(x_{i},x_{j})$ is defined as

\begin{equation}
\varphi _{p}(x_{i},x_{j})=\mu(x_{i},x_{j})\underbrace{\sum_{m=1}^{K}w^{(m)}k^{(m)}(f_{i},f_{j})}_{k(f_{i},f_{j})}
\end{equation}

where $f_{i}$ and $f_{j}$ are feature vectors for pixels $i$ and $j$ in a feature space, $w^{m}$ linear combination weights. $\mu$ is a label compatibility function. $k^{(m)}$ are $m$ Gaussian kernels. For multi-class segmentation, three contrast-sensitive components are defined in terms of intensity vectors $I_i$ and $I_j$ in RGB channels, positions $p_i$ and $p_j$, and the extra features $C_i$ and $C_j$:

\begin{equation}
\begin{split}
k(f_{i},f_{j})=w^{(1)}exp(-\frac{\left | p_{i}-p_{j} \right |^{2}}{2\theta ^{2}_{\alpha }} 
-\frac{\left | I_{i}-I_{j} \right |^{2}}{2\theta ^{2}_{\beta  }}\\
-\frac{dist(C_{i}-C_{j})^{2}}{2\theta ^{2}_{\gamma }}) 
+w^{(2)}exp(-\frac{\left | p_{i}-p_{j} \right |^{2}}{2\theta ^{2}_{\delta  }})
\end{split}
\label{eq9}
\end{equation} 

In Equation \ref{eq9}, the first term is appearance kernel, based on the observation that pixels and the extra features nearby with similar colors are likely in the same class. $\theta_{\alpha}$, $\theta_{\beta}$ and $\theta_{\gamma}$ control the spatial distance and the similarity of intensity and the extra feature. The second term is smoothness kernel, removing small isolate regions. The extra features $C_i$ and $C_j$ can be selected from ToF, horizontal disparity, variance, the angle that local normal makes with the inferred gravity direction, and HVA channels at pixels $i$ and $j$. The $dist$ function is the Euclidean distance between two vectors. Similar to \cite{krahenbuhl2011efficient}, we use grid search on our validation set for the parameters $w^{(1)}$, $\theta_{\alpha}$, $\theta_{\beta}$ and $\theta_{\gamma}$, and assign 1.0 to $w^{(2)}$ and $\theta_{\delta}$ in practice.

\begin{figure}
	\centering
	\includegraphics[width=1\linewidth]{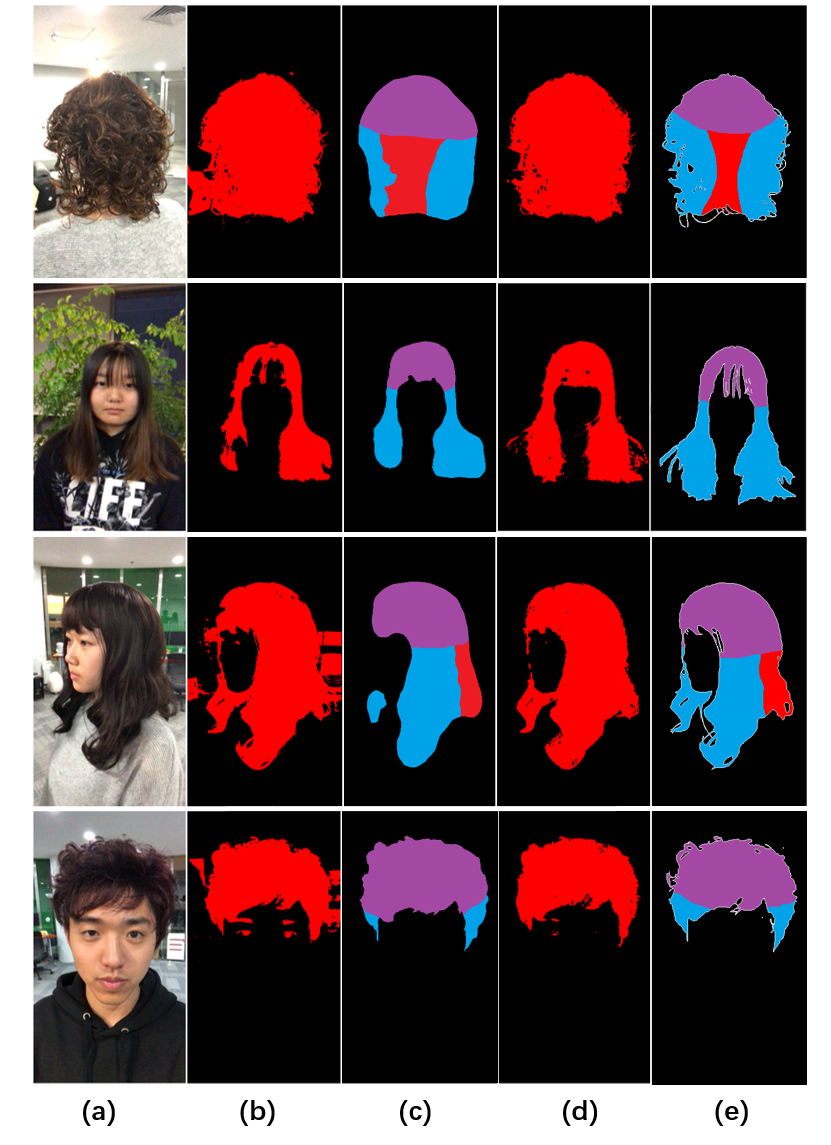}
	\caption{Visual comparisons of our technique vs. CRFs \cite{krahenbuhl2011efficient} and DeepLabv3+ \cite{chen2018encoder}.}  
	\label{fig:result2}  
\end{figure}

\section{Experimental Results}
We first qualitatively validate our technique on the ToF hair dataset and compare the performance with state-of-the-art techniques DeepLabv3+ in \cite{chen2018encoder} and CRFs in \cite{krahenbuhl2011efficient}.

We conduct hair segmentation experiments for various hairstyles on different scenes. Our dataset comprises 20k+ head images of 30 subjects. We manually annotate 8k images, 5 labels for segmentation and 5 labels for hair directional map. The segmentation labels include background, facial region and hair regions of top, back and two sides. Hair direction labels are leftward, rightward, horizontal and longitudinal. Further, we categorize these annotated images into three groups: 6k for training, 1.2k for testing and 0.8k for validation. All our experiments are carried out on the public platform TensorFlow \cite{tensorflow} on a PC machine with two graphics cards of TiTan 1080 Ti.

\begin{table}[]
\caption{Evaluation of our ToF-HairNet components and state-of-the-art techniques on mIoU metric at 50k training times (D3 denotes DeepLabv3+)}
\centering
\begin{tabular}{l|l|l|ll}
\cline{1-4}
\multicolumn{1}{c|}{Method} & \multicolumn{1}{c|}{mIoU} & \multicolumn{1}{c|}{Method} & \multicolumn{1}{c}{mIoU} &  \\ \cline{1-4}
D3\cite{chen2018encoder}  & 0.69 & FIGARO\cite{2016ICIP} & 0.60 &  \\
D3+ToF(origin) & 0.73  & PSPNet\cite{2017CVPRzhao} & 0.69 & \\
D3+Variance & 0.75  & FCN\cite{long2015fully} & 0.63 &  \\
D3+HVA  & 0.77  & Autohair\cite{chai2016autohair}  & 0.65 &  \\
D3+direction & 0.72  & D-CNN\cite{wang2018depth}  & 0.71 &  \\
\textbf{ToF-HairNet} & 0.81  & FuseNet\cite{hazirbas2016fusenet} & 0.67 &  \\ \cline{1-4}
\end{tabular}
\label{tab:miou_NN}
\end{table}

\begin{table}[]
\caption{Component evaluation of CRFs+X for hair segmentation refinement on IoU metric} 
\centering
\begin{tabular}{l|l|l|ll}
\cline{1-4}
\multicolumn{1}{c|}{Method} & \multicolumn{1}{c|}{IoU} & \multicolumn{1}{c|}{Method} & \multicolumn{1}{c}{IoU} &  \\ \cline{1-4}
CRFs\cite{krahenbuhl2011efficient}  & 0.830 & CRFs+ToF & 0.841 &  \\
CRFs+Disparity & 0.840  & CRFs+Normal & 0.833 & \\
CRFs+Variance(noise) & 0.851  & \textbf{CRFs+HVA} & 0.860 &  \\
\cline{1-4}
\end{tabular}
\label{tab:miou_CRF}
\end{table}

We build our network based on the 65-block Xception network, and fine-tune its parameters initiated on ImageNet dataset \cite{imagenet} with similar training protocol in \cite{chen2018encoder}. In particular, we set the learning rate to 0.0001, crop-size to 513$\times$513$\times$3, training step to 50k times, output stride to 16 for encoder and 4 for decoder. We select the random mode for cropping position, flip, scale, and exposure to perform average operation on the annotated RGB images. Exploiting the parameters fine-tuned, we train hair segmentation on our ToF-HairNet following the details in \ref{section:network}, and output hair directional map, the segmentation and its labeling. We use Adam optimizer~\cite{kingma2014adam},and set $beta_{1}=0.9$,$beta_{2}=0.99$. The batch size is set to 4, the learning rate drops 10 times every 100 epoch, and the early stop training strategy is used. Other parameters such as batch normalization are set as default parameters.For the parameter of CRFs, $\theta_{\alpha}$,$\theta_{\beta}$,$\theta_{\gamma}$,$\theta_{\delta}$ are set as 25,25,35,45.

Our experimental results on the challenging scenes, e.g, hair color similar to the background, hair on the temples and hair highlighted, are shown in Figure \ref{fig:tensor} in comparison to those using D-CNN \cite{wang2018depth}, CRFs \cite{krahenbuhl2011efficient}, and DeepLabv3+ \cite{chen2018encoder}. We show that RGB based methods result in missed hair boundary and details merely dependent upon intensity variations and spatial positions. As additional cues, each channel of HVA, in particular variance variation on ToF images is capable of more reliable segmentation and labeling for specific surfaces such as hair. 

Further, we conduct our experiments on various hairstyles compared with CRFs \cite{krahenbuhl2011efficient} and DeepLabv3+ \cite{chen2018encoder}. Figure \ref{fig:result2} shows sample results of 4 hairstyles. Our hair masks are significantly more accurate than two state-of-the-art techniques, especially for the fluffy and curly hairstyles.

\begin{figure}
	\centering
	\includegraphics[width=1\linewidth]{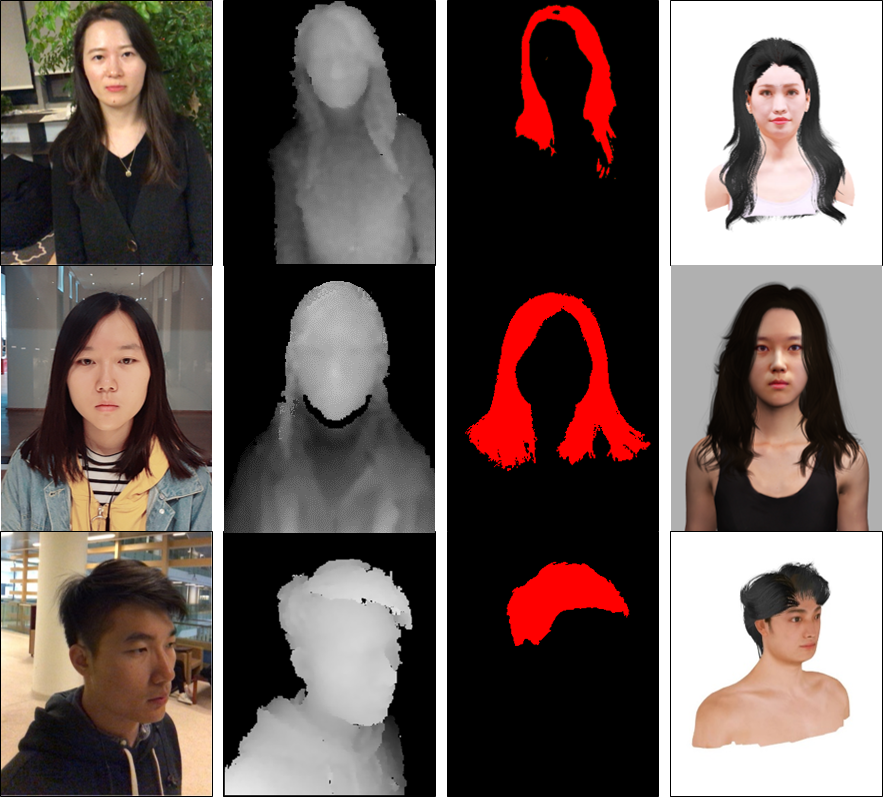}
    \caption{An application example in hair style identification. }
	\label{fig:application}  
\end{figure}

Quantitatively, we evaluate the components of our ToF-HairNet and other networks on mean intersection over union (mIoU) metric\cite{mIoU}. The IoU measures the number of pixels between the ground truth and segmentation masks divided by the total pixels present across both masks. We calculate IoU score for each label and average the IoU score over all annotated labels to obtain mIoU of our network when we combine each component separately and combine all on base of DeepLabv3+ \cite{chen2018encoder}. Our mIoU metric in Table \ref{tab:miou_NN} shows the mean IoU at 50k training times. As shown in left side of Table \ref{tab:miou_NN}, ToF images, components computed from the ToF images, hair gradient and directional maps contribute significantly to the mIoU score of our network. 
We believe that ToF images can add additional constraint information to the neural network, which will effectively improve the detection accuracy. As Table \ref{tab:miou_NN} shows, Depth constraint is 4$\%$ higher than using RGB alone as input. At the same time, more analysis are focused in Variance Map. We can find that the variance can effectively improve the segmentation accuracy, which is $2\%$ higher than the single addition of depth information. Like \cite{gupta2014learning}, we modify the depth channel to 3 channels of HVA space information, which can greatly improve the segmentation accuracy. Experiments can show that the depth noise is helpful to improve the accuracy of hair segmentation.

After adding direction result supervision, the segmentation effect has also improved. It can be considered that multi-tasking has further improved the representation of the network. Adding gradient information (Sobel operator) alone will not greatly improve the result. However, it is found in experiments that adding a gradient constraint will help improve the convergence of the network without increasing much computational complexity.
We conduct experiments compared with state-of-the-art techniques \cite{2016ICIP, chai2016autohair,  chen2018encoder,2017CVPRzhao,long2015fully} using the RGB images, and \cite{wang2018depth,hazirbas2016fusenet} using the RGBD images in our hair dataset. The results in Table \ref{tab:miou_NN} show that our method is outstanding in those works. 

Alternatively, we evaluate the components of our refinement segmentation by combining CRFs with null, ToF, each channel of HVA, and three channels of HVA only on hair part. The experimental results in Table \ref{tab:miou_CRF} demonstrate that ToF images and the components extracted from ToF images can highly improve hair segmentation. Our refinement process is able to remove the hairy areas such as eyebrows segmented along with hair regions in the initial segmentation (shown in Figure \ref{fig:result2}), but it may fail when hair strands on the forehead are dense and long to reach down to the eyebrows.

\begin{figure}
	\centering
	\includegraphics[width=1\linewidth]{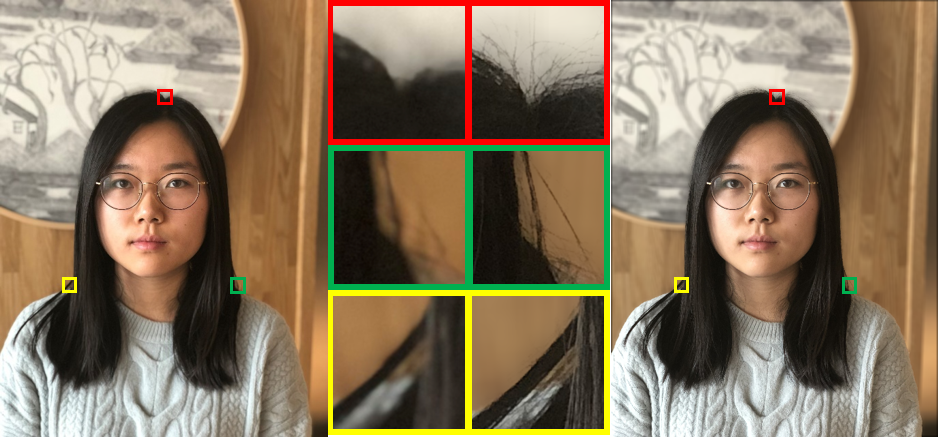}
    \caption{An application example in image enhancement. A photo taken with iPhone X portrait mode (a) and its enhancement with our hair mask (b). }
	\label{fig:matting}  
\end{figure}

We finally show our segmentation results in Figure \ref{fig:application} to identify their corresponding styles and use CG to re-render both hair and face to create more realistic avatars, and in Figure \ref{fig:matting} to produce hair mattes for image enhancement using the technique in \cite{he2011global}. We refer the reviews to the appendix for many additional results.

\section{Conclusion}
While noise caused by scattering and inter-reflection in ToF has been viewed adversarial to 3D scanning, we have exploited this unique property to assist hair segmentation. Our approach however is limited by low depth discrepancy on existing ToF sensors: if the camera is far away from the subject, it'd be challenging to conduct the variance estimation as the subject (along with hair) appears at a single depth layer. This problem may be mitigated with emerging high precision ToF sensors. Robust segmentation of hair has numerous applications. In the paper, we have demonstrated synthetic depth-of-field imaging with high quality hair matte as a direct result obtained from segmentation. An immediate future direction is to extend our approach to multi-view settings for 3D hair reconstruction. Most existing approaches such as Structure-from-Motion model hair as a coarse mesh and the visual realism is far from satisfactory. In contrast, in photorealistic rendering, hair is often modeled as strands. Our ToF solution may provide a viable tool for generating such hair strand models: our noise analysis can be potentially used to measure hair fiber density and our gradient estimation may directly guide the construction of strands.

\ifCLASSOPTIONcaptionsoff
  \newpage
\fi



%
%
{\small
\bibliographystyle{IEEEtran}
\bibliography{egbib}
}
%


%







\end{document}